\documentclass[10pt,twocolumn,letterpaper]{article}

\usepackage{cvpr} 

\usepackage[dvipsnames]{xcolor}

\usepackage{svg}
\usepackage{multirow}
\usepackage{array,multirow,graphicx}
\usepackage{float}
\usepackage{stfloats}
\usepackage{balance}
\usepackage[accsupp]{axessibility}
\definecolor{cvprblue}{rgb}{0.21,0.49,0.74}
\usepackage[pagebackref,breaklinks,colorlinks,citecolor=cvprblue]{hyperref}

\title{Fusion Transformer with Object Mask Guidance for Image Forgery Analysis \vspace{-4pt}}

\author{
Dimitrios Karageorgiou$^1$ \quad Giorgos Kordopatis-Zilos$^2$ \quad Symeon Papadopoulos$^1$  \vspace{2pt} \\
$^1$Information Technologies Institute, CERTH\\
$^2$VRG, FEE, Czech Technical University in Prague \\
{\tt\small \{dkarageo,papadop\}@iti.gr}  \quad
{\tt\small kordogeo@fel.cvut.cz} \vspace{-3pt}
}

\begin{document}
\maketitle
\begin{abstract}
\vspace{-7pt} 
In this work, we introduce \textit{OMG-Fuser}, a fusion transformer-based network designed to extract information from various forensic signals to enable robust image forgery detection and localization. Our approach can operate with an arbitrary number of forensic signals and leverages object information for their analysis -- unlike previous methods that rely on fusion schemes with few signals and often disregard image semantics. To this end, we design a forensic signal stream composed of a transformer guided by an object attention mechanism, associating patches that depict the same objects. In that way, we incorporate object-level information from the image. Each forensic signal is processed by a different stream that adapts to its peculiarities. A token fusion transformer efficiently aggregates the outputs of an arbitrary number of network streams and generates a fused representation for each image patch. We assess two fusion variants on top of the proposed approach: (i) score-level fusion that fuses the outputs of multiple image forensics algorithms and (ii) feature-level fusion that fuses low-level forensic traces directly. Both variants exceed state-of-the-art performance on seven datasets for image forgery detection and localization, with a relative average improvement of 12.1\% and 20.4\% in terms of F1. Our model is robust against traditional and novel forgery attacks and can be expanded with new signals without training from scratch. Our code is publicly available at: \url{https://github.com/mever-team/omgfuser}
\vspace{-10pt}
\end{abstract}    
\section{Introduction}\label{sec:introduction}

\setlength{\textfloatsep}{3pt}{
\begin{figure}
  \centering
  \includesvg[width=0.47\textwidth]{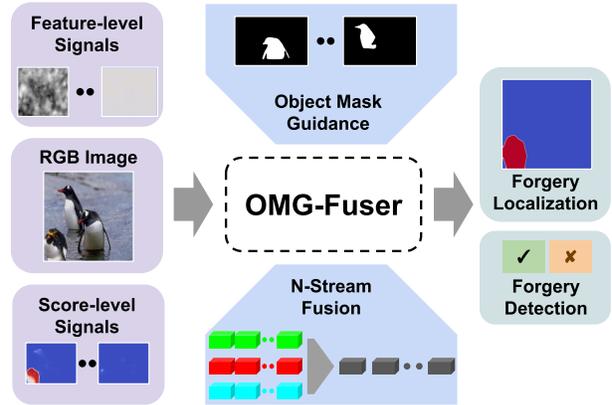}
  \vspace{-8pt}
  \caption{OMG-Fuser combines an arbitrary number of heterogenous forensic signals for robust image forgery analysis guided by the image semantics.
  }
  \label{fig:summary}
\end{figure}
}

Digital image forgery~\cite{kaur2023image} is increasingly becoming more accessible and efficient due to the pervasive availability of sophisticated image editing algorithms as part of free or low-cost image editing applications for desktops and smartphones. Notably, multiple works~\cite{schetinger2017humans, nightingale2017can} have found that the ability of humans to detect forged images hardly exceeds the performance of random guessing, especially when the forgery is of decent or high quality. Despite the significant advances in the field of image forensics~\cite{kaur2023image, verdoliva2020media, zanardelli2023image}, existing detection methods are greatly challenged when tested in the wild without strong assumptions about the processing history of an image \cite{zanardelli2023image, wu2022robust}. To this end, the main focus of our work is to robustly detect the forged regions within images, along with an overall decision for the image, formally defined as \textit{image forgery localization and detection} respectively~\cite{zheng2019survey}. \looseness=-1

A common practice for image forensics analysts is to utilize various image forensics methods to increase the chances of detecting forgery. However, in such cases, analysts need to judge based on their experience what tool to trust, an issue further exacerbated by the recent deep learning approaches that operate as black boxes. This has given rise to approaches that fuse multiple forensic signals to capture more robust forensic clues. Such approaches can be classified into two main categories~\cite{phan2022comparative}. (i) \textit{Feature-level fusion}, which fuses low-level features extracted from the input image that usually represent different domains \cite{zhou2018learning, kwon2022learning, guillaro2023trufor}. 
However, these purpose-built feature fusion architectures are only effective in detecting specific types of forgery. (ii) \textit{Score-level fusion}~\cite{phan2022comparative}, which combines the outputs of multiple different forensics methods into a single output \cite{fontani2013framework, phan2022comparative, charitidis2021operation}. However, theoretical limitations or the employed training approaches prevent current score-level fusion approaches from effectively combining the outputs of recent deep learning algorithms. Additionally, while image semantics could greatly help the fusion process, the impact of such a direction is underexplored in the  literature~\cite{wang2022objectformer}. \looseness=-1

To this end, we propose the \textit{Object Mask-Guided Fusion Transformer (OMG-Fuser)}, capable of capturing image forensic traces from an arbitrary number of input signals by leveraging the image semantics and fusing them in multiple processing stages into a single robust forgery detection and localization output. Our fusion network comprises two main modules: the \emph{Forensic Signal Streams} and the \emph{Token-Fusion Module}. In the former, every input signal is propagated through a different stream to capture its unique traces. Our main novelty in this stage includes the \emph{Object Guided Attention} mechanism that exploits external instance segmentation masks to drive the extraction process to attend only to image regions that depict the same objects so as to generate comprehensive object-level representations. Any pretrained instance segmentation model can be employed~\cite{gu2022review}, and we utilize a recently proposed class-agnostic model~\cite{kirillov2023segment}. The latter module is responsible for the fusion of any arbitrary number of forensic signals. The \emph{Token-Fusion Transformer (TFT)} combines the various representations of an image region generated by the forensic signal streams into a single representation via several transformer blocks. The fused outputs are analysed by the \emph{Long-range Dependencies Transformer (LDT)} that captures the relations between the representations of the different image regions. For training the proposed network, we propose the \emph{Stream Drop} augmentation. This randomly discards some network streams during training to prevent the network from over-attending on specific forensic signals. Our proposed approach is employed both for feature and score level fusion, combining RGB information with two and five forensic signals, respectively. We demonstrate its effectiveness on five popular datasets, compared with several handcrafted and deep learning state-of-the-art approaches. Moreover, we demonstrate its robustness against common perturbations and recent neural filters. An overview of the approach is presented in \cref{fig:summary}.

In summary, our contributions include the following:
\begin{itemize}
    \item We propose a fusion transformer for robust image forgery detection and localization that analyzes an arbitrary number of image forensic signals based on image semantics.
    \item We introduce the object guided attention that uses object-level information to drive the attention process.
    \item We design a token-fusion transformer that combines an arbitrary number of patch tokens into a single comprehensive representation for each image region.
    \item We introduce the stochastic augmentation process, named stream drop, for avoiding over-attending on particular streams while training multi-stream networks.
    \item We improve the state-of-the-art by 12.1\% and 20.4\% F1 on image forgery detection and localization, respectively.
\end{itemize}
\setlength{\textfloatsep}{3pt}
\setlength{\dbltextfloatsep}{3pt}
\setlength{\floatsep}{3pt}
\setlength{\dblfloatsep}{3pt}
\setlength{\abovecaptionskip}{2pt}
\setlength{\belowcaptionskip}{2pt}
\section{Related Work}\label{sec:related_work}

\subsection{Image Forgery Detection and Localization}

Image editing operations introduce subtle but detectable traces \cite{verdoliva2020media}. Most early works in the field of image forensics focused on detecting traces of a single type of forgery using handcrafted signal processing operations. These include, for instance, disturbances related to the Color Filter Array (CFA)~\cite{popescu2005exposing, ferrara2012image, bammey2021image} or noise inconsistencies using the popular PRNU pattern~\cite{lukas2006digital, chierchia2011prnu, zhang2023prnu}. Others employ filtering and frequency analysis \cite{cozzolino2015splicebuster, fridrich2012rich} or extract and analyze noise residuals \cite{mahdian2009using}. Also, many works focus on detecting artifacts introduced by lossy compression algorithms, such as JPEG \cite{lin2009fast, bianchi2011improved, iakovidou2018content, nikoukhah2021zero, farid2009exposing}.

\begin{figure*}
  \centering
  \vspace{-7pt}
  \includesvg[width=0.93\textwidth]{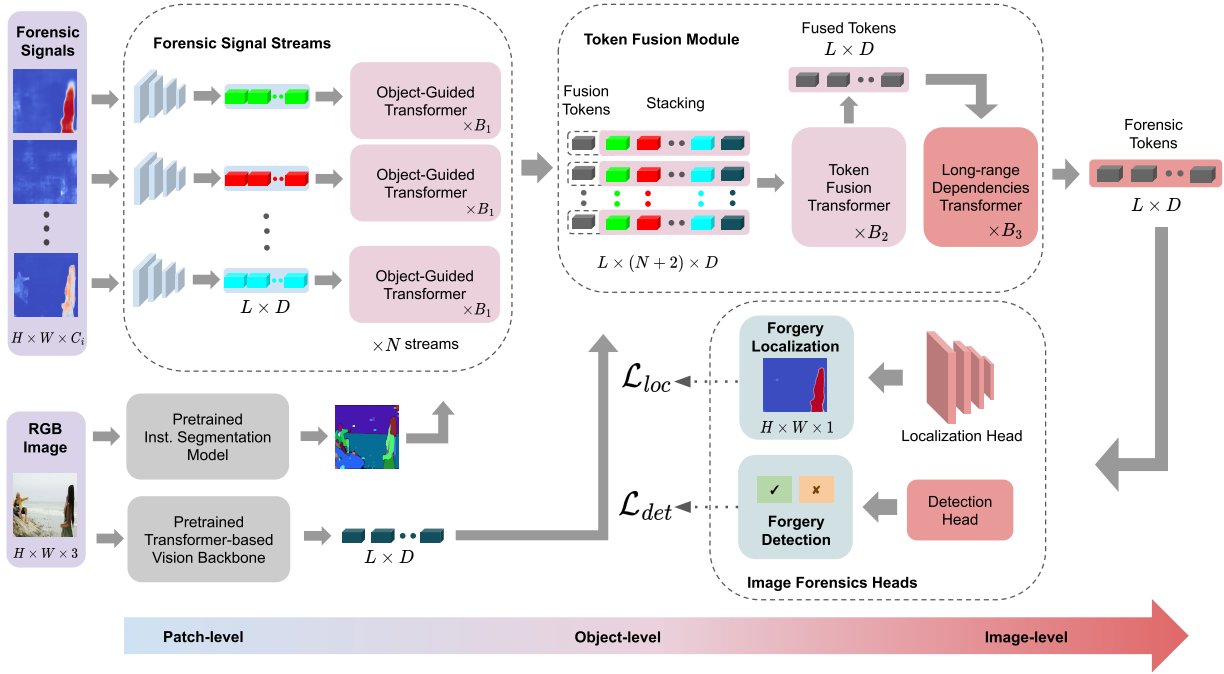}
  \caption{Overview of OMG-Fuser. Forensic signals are fused into a robust forgery localization mask and detection score. To achieve that, it combines information from the RGB image and its instance segmentation maps. Each forensic signal and the RGB image are first processed by separate network streams through independent Object-Guided Transformers. Then, the proposed Token Fusion Module fuses the different streams, leading to features with a progressively increasing level of information granularity, from patch-level (in the early stages) to object-level (in intermediate stages) and to image-level (in the final stages). A localization and a detection head process the extracted forensic tokens to generate the final outputs.}
  \vspace{-2pt}
  \label{fig:approach_overview}
\end{figure*}

Recent advances in deep learning have reshaped the field of image forensics. Instead of handcrafted features that capture a narrow range of artifacts and are prone to naive post-processing operations, recent methods employ deep neural networks that learn to capture more robust forensic traces for detecting forged regions. Such  methods~\cite{cozzolino2019noiseprint, kwon2022learning, hua2023learning, guillaro2023trufor} dominate the state-of-the-art, outperforming the previous approaches usually by a large margin. Several CNN-based architectures have been proposed, targeting one \cite{liu2021two, agarwal2020efficient, elaskily2020novel} or more \cite{cozzolino2019noiseprint, rajini2019image, zhang2016image} types of forgery, while more recently, LSTM-based \cite{bappy2019hybrid} and transformer-based \cite{liu2023tbformer, guillaro2023trufor, sun2022edge} architectures have emerged. Even though deep learning approaches can capture more complex artifacts, the employed network architectures and training procedures still limit their detection capability to a narrow range of forensic artifacts, \eg focusing only on noise anomalies ~\cite{wu2019mantra,hu2020span} or capturing JPEG compression artifacts~\cite{kwon2022learning}. 

\subsection{Fusion Approaches}
According to Phan-Ho \etal \cite{phan2022comparative}, fusion approaches in the image forensics literature can be classified into two broad categories: feature-level and score-level fusion.

\textbf{Feature-level fusion approaches} combine feature representations incorporating information from different domains into a comprehensive one that can be employed for detecting a broader range of forgery cases. A seminal work in the field by Zhou et al.~\cite{zhou2018learning} proposed a two-stream network architecture: one network stream acts on the RGB image to detect forgery artifacts, and the other captures noise-related information by processing the output of SRM filters. Many state-of-the-art methods have employed such a feature fusion approach, each using different inputs and building purpose-specific fusion architectures for better capturing traces such as compression-related artifacts~\cite{kwon2022learning, wang2022objectformer}, noise disturbances~\cite{dong2022mvss, guillaro2023trufor} or anomalous edges~\cite{dong2022mvss}. However, while current feature fusion approaches significantly boost detection performance in the target forgery type, they often fail when deployed in the wild \cite{zanardelli2023image, wu2022robust} and are designed for fusing a small number of specific input signals.

\textbf{Score-level fusion approaches} combine the outputs of multiple forgery detection and localization algorithms to leverage the benefits of each one into a single output. A widely adopted practice in the field is to utilize statistical approaches, such as the Dempster-Shafer Theory (DST) \cite{fontani2013framework, ferrara2015unsupervised} or the Bayes' Theorem \cite{korus2016multi, korus2016multi2} to derive a unified prediction from multiple outputs of different image forensics algorithms. However, these require strong theoretical guarantees regarding the compatibility relations and statistical independence between the fused methods, requirements that are very tough to meet in recent deep-learning networks that operate as black boxes. Hence, they are not compatible with the current state-of-the-art algorithms. As a more general solution, learning approaches have recently been proposed for fusing the outputs of multiple image forensic algorithms \cite{charitidis2021operation, ferreira2016behavior, siopi2023multi}. However, they still rely on forensic signals generated by handcrafted signal processing operations as input and lack mechanisms for effectively leveraging image semantics and for preventing over-attendance on the best signals, all of which make them unsuitable for fusing state-of-the-art deep learning-based forensic signals. 

Finally, recent works \cite{wang2022objectformer, zhuo2022self} highlight the beneficial impact of utilizing semantic relations within images for forgery detection. They design trainable network components tailored to capture information related to image semantics. However, learning to incorporate information about all the possible objects encountered in-the-wild into such a network demands huge amounts of training data. To the best of our knowledge, we are the first in the field of image forensics to employ a pre-trained instance segmentation model to introduce object-level information into the attention mechanism, which requires no further training on large-scale datasets. Furthermore, there is no similar prior work in the image forensics domain that builds a transformer-based architecture that can address feature-level and score-level fusion -- prior related works~\cite{wang2022objectformer,zhou2018learning,guillaro2023trufor,charitidis2021operation,siopi2023multi} focus on one of the two categories, and usually are limited in terms of the number of input signals that can be combined. 
\section{Approach Overview}\label{sec:approach}

This section presents our proposed \textbf{O}bject \textbf{M}ask \textbf{G}uided \textbf{Fus}ion Transform\textbf{er} (OMG-Fuser), a network architecture for fusing multiple image forensic signals by leveraging image semantics. \cref{fig:approach_overview} illustrates the proposed architecture.

\subsection{Problem formulation}

Given an RGB image $x^{rgb} \in \mathbb{R}^{H \times W \times 3}$ and a number of $N$ forensic signals $x^{sig}_{i} \in \mathbb{R}^{H \times W \times C_{i}}, i \in \{1... N\}$, the goal is the OMG-Fuser network to predict a pixel-level forgery localization mask $\hat{y}^{loc} \in (0, 1)^{H \times W \times 1}$ and an image-level forgery detection score $\hat{y}^{det} \in (0, 1)$. $H$ and $W$ denote the height and width of the input image, respectively, while $C_{i}$ denotes the number of channels of the $i^{\text{th}}$ forensic signal. 

\subsection{Forensic Signal Streams} \label{sec:fcs}

Each of the $N$ forensic signals $x^{sig}_{i}$ capture different artifact types. To capture their characteristic elements, we pass each signal through a different network stream, denoted as \textit{Forensic Signal Stream (FSS)}. It consists of two main components, i.e., patch representations and an object-guided transformer.

\textbf{Patch representations:} Given that FSS is built upon transformer blocks, we need to convert the 2D input signals to a set of tokens. Following the common practice in the field~\cite{wang2022objectformer, xiao2021early,dosovitskiy2020image}, a convolutional neural network (CNN) with four strided layers is applied to the input. Its output $\bar{x}^{sig}_{i} \in R^{H' \times W' \times D} $ consists of feature representations extracted from the input forensic signals corresponding to specific image patches. $H' = H / p$ and $W' = W / p$ are the output's spatial dimensions, $p$ denotes the size of the patches in the initial image, and $D$ is the latent dimensionality. Finally, we reshape $\bar{x}^{sig}_{i}$ from $R^{H' \times W' \times D}$ to $R^{L \times D}$, where $L = H' W'$, to form our token sequence, which is our patch-level representations.

\textbf{Object-Guided Transformer (OGT):} This is the key component to infuse object-level information in the patch representations, effectively converting them to object-level representations. Inspired by Vision Transformer (ViT) \cite{dosovitskiy2020image} and Masked-Attention \cite{cheng2022masked}, we design a process that utilizes instance segmentation maps generated by an instance segmentation model to restrict the transformer's attention to tokens belonging to the same objects. 

Given an input image, a number of 2D instance segmentation maps $x^{seg}_j \in \{0, 1\}^{H \times W \times 1}, j \in \{1...K\}$ are generated corresponding to a set of objects $\mathcal{S}=\{o_1, o_2, ..., o_K\}$ depicted in the image. 
Using these instance segmentation maps, we can extract a subset $\mathcal{S}_u \subset \mathcal{S}$ for each image patch $u$, containing all objects depicted in the corresponding $p \times p$ area in the original image. An object belongs to a patch if at least one pixel of the patch has been annotated with the corresponding object label. Hence, we define the object-guided attention mask $M \in \{-\infty, 0\}^{L \times L}$ based on the patches that contain the same objects and in order to be used during the attention calculation in the next operation. The mask generation can be formulated as
\begin{equation}
    M_{(u,v)} = 
        \begin{cases}
            0 &\quad\text{if } \mathcal{S}_u \cap \mathcal{S}_v \neq  \emptyset \\
            -\infty & \quad\text{otherwise} \\
        \end{cases},
\end{equation}
where $u,v \in \{1...L\}$ denote two arbitrary image patches. In that way, during the attention calculation, image patches that depict the same objects will attend to each other, while others will be ignored.

It is noteworthy that instance segmentation models do not always generate maps covering all image pixels. Pixels that have not been annotated with any instance labels are considered as background and are annotated with a corresponding label. In that way, we allow the network to focus on regions with no detected instances, which is especially useful in forgery cases where entire objects have been removed from the original image, i.e., inpainting. The attention regions defined by the object-guided attention mask for a sample case are displayed in \cref{fig:igam}.

\setlength{\textfloatsep}{5pt}{
\begin{figure}
  \centering
  \includesvg[width=0.47\textwidth]{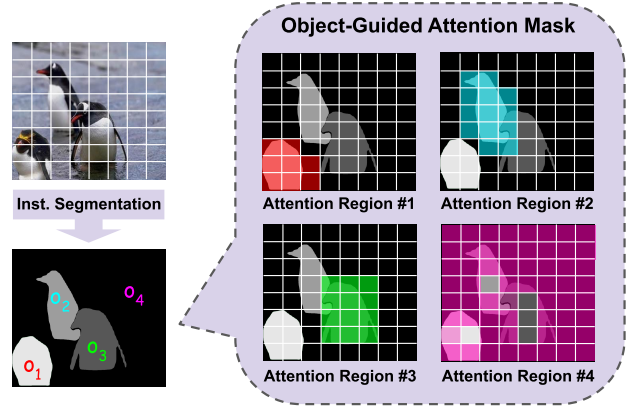}
  \caption{Object-Guided Attention Mask: Limits the attention of the transformer only between patches that depict the same objects. The four attention regions defined by the mask for an example image are depicted to the right. The background is considered as another object. For illustration purposes, the number of patches on both axes has been limited to eight.}
  \label{fig:igam}
\end{figure}
}

\emph{Object-Guided Attention (OGA)}:
We extend the self-attention mechanism \cite{vaswani2017attention} with our object-guided attention masks to guide tokens to attend only to tokens belonging to the same objects and ultimately derive object-level representations. Let an arbitrary token sequence $z \in \mathbb{R}^{L \times D}$ and its queries, keys and values $Q, K, V \in \mathbb{R}^{L \times D_h}$ be its projections for the self-attention. $D_h$ denotes the dimensionality of the latent representation of the attention process. Then, the OGA calculation can be done as follows
\begin{eqnarray}
Q=zW_q,\quad K=zW_k,\quad V=zW_v \\
OGA(z) = \sigma((QK^T + M) / \sqrt{D_h}) V,
\end{eqnarray}
where $W_q,W_k,W_v \in \mathbb{R}^{D \times D_h}$ are the projection matrices, and $\sigma$ denotes the softmax function. In that way, our masks force this process to attend only to token pairs with zero-value elements in the mask $M$.

In our implementation, we follow the multi-head version of self-attention ~\cite{vaswani2017attention}. To build our OGT, we use transformer blocks similar to ViT~ \cite{dosovitskiy2020image}. We first apply our OGA mechanism, followed by a Feed Forward network that acts on each token separately, consisting of an MLP with GELU activation~\cite{hendrycks2016gaussian}. Both operations are applied with a residual connection and a Layer Normalization~\cite{ba2016layer} before their application. Finally, each forensic signal $x^{sig}_i$ is processed by a dedicated stream that extracts patch representations that are processed by an OGT, consisting of $B_1$ transformer blocks, which outputs object-level representations $z^{sig}_i \in R^{L \times D}$ for the image patches.

 \textbf{RGB Stream:} To encode and analyze the RGB image, we opt for using transformer-based backbone networks pretrained on large-scale image collections~\cite{han2022survey, vaswani2017attention} to capture relevant information in the RGB domain. Specifically, the $x^{rgb}$ image is propagated through a pretrained vision backbone network \cite{oquab2023dinov2}, which composes our \textit{RGB Stream}, to extract representations for the image regions. The output of this processing is denoted as $z^{rgb} \in R^{L \times D}$.

\subsection{Token Fusion Module} \label{sec:token_fusion_module}

Up to this point, the network has processed each of the $N$ forensic signals and the RGB image using a separate stream, in a total of $N+1$ network streams. Practically, we have $N+1$ tokens for each image patch; hence, our goal is first to fuse all this information to a single representation for each patch and then capture the relation between the fused patch tokens. To this end, we propose the \textit{Token Fusion Module (TFM)}, which comprises two main components: (i) the \textit{Token Fusion Transformer (TFT)} for satisfying the former requirement, and (ii) the \textit{Long-range Dependencies Transformer (LDT)}, for satisfying the later requirement. In that way, we transform the object-level features to image-level ones, which can be utilized for various downstream image forensics tasks. 

\textbf{Token Fusion Transformer (TFT):} To shape the patch tokens generated from the RGB stream and the FSS into a form that our transformer can process, we stack all tokens together as follows:
\begin{equation} \label{eq:streams_concat}
    z = [z^{sig}_1; z^{sig}_2;...;z^{sig}_N;z^{rgb}] \quad \in \mathbb{R}^{(N+1) \times L \times D}
\end{equation}
where $z$ is the patch token tensor and $[\cdot; \cdot]$ denotes the stacking function on the outermost dimension. Hence, the $N+1$ tokens originating from the same image patch but from different streams are organized together. To fuse patch tokens derived from the same image patch, we follow the common practice in transformer literature~\cite{vaswani2017attention,dosovitskiy2020image}: in particular, we employ a learnable \emph{fusion token} $z^{ft} \in \mathbb{R}^{L \times D}$ repeated and stacked with the other patch tokens for each image patch. The goal is to refine the fusion tokens through the attention process of the TFT so as to incorporate relevant information from the $N+1$ tokens generated by the different streams. For the implementation of the TFT, we use $B_2$ transformer blocks~\cite{dosovitskiy2020image}. Also, the TFT attention is applied to the outermost dimension of tensor $z$, which is considered the token sequence dimension. This can be formulated as follows:
\begin{equation}
     [\bar{z}^{ft}; \bar{z}] = TFT([z^{ft}; z]),
\end{equation}
where $\bar{z}^{ft} \in \mathbb{R}^{L \times D}$ are the \emph{fused tokens}. 
After TFT, only the $\bar{z}^{ft}$ is considered for further processing, and the refined tensor $\bar{z}$ is discarded.

\textbf{Long-range Dependencies Transformer (LDT):} The fused tokens at the TFT output incorporate the forensic information coming out of all the $N+1$ streams. However, the information is aggregated so far only at the object level. Hence, we introduce the LDT component to take into account the forensic information at a global level and suppress wrongly captured inconsistencies while highlighting subtle but crucial traces. The LDT employs the transformer architecture~\cite{dosovitskiy2020image} to refine the fused tokens $\bar{z}^{ft}$ into \textit{forensic tokens} $z^{for} \in \mathbb{R}^{L \times D}$, by enabling the attention between all the $L$ tokens. In that way, forensic tokens incorporate image-level information that derives from the relations between them, capturing long-range dependencies. For the implementation of LDT, a total of $B_3$ transformer blocks~\cite{dosovitskiy2020image} are used.

\subsection{Output Generation}\label{sec:forensics_tasks}

To generate the outputs for the two target forensics tasks, we employ two specialized heads to further process the forensic tokens $z^{for}$. The first is the localization head, consisting of five transposed convolution layers, that generates the forgery localization output $\hat{y}^{loc}$. The second is the detection head, employing a four-block transformer~\cite{dosovitskiy2020image} and a fully connected layer to predict the detection score $\hat{y}^{det}$.

\subsection{Training process} \label{sec:training}

\textbf{Loss functions: } To equally weight the error for the pristine and tampered regions, despite their expected different size on each sample, we employ the balanced Binary Cross Entropy (bBCE) loss $\mathcal{L}_{bBCE}$, for equalizing the contribution of both regions in the loss. We compute the bBCE by averaging the binary cross entropy loss computed separately for each of the two regions, i.e., pristine and tampered. Furthermore, to generate cleaner forgery localization masks, we also employ the Dice loss \cite{milletari2016v, guillaro2023trufor} $\mathcal{L}_{dice}$ into the localization loss $\mathcal{L}_{loc}$. Regarding forgery detection loss $\mathcal{L}_{det}$, we employ the BCE loss. Thus, with $a$, $b$ and $c$ being weighting hyperparameters, the total training objective becomes \looseness=-1
\begin{equation}
    \mathcal{L} = \underbrace{(a \cdot \mathcal{L}_{bBCE} + b \cdot \mathcal{L}_{dice})}_{\mathcal{L}_{loc}} + c \cdot \mathcal{L}_{det}.
\end{equation}

\begingroup
\setlength{\tabcolsep}{5pt}
\begin{table*}
    \centering
    \scalebox{0.81}{
    \begin{tabular}{ c l | cc cc cc cc cc cc cc | cc }
    \toprule
    \multicolumn{2}{c}{\multirow{2}{*}{\textbf{Approach}}}& \multicolumn{2}{c}{\textbf{CASIAv1+}} & \multicolumn{2}{c}{\textbf{Columbia}} & \multicolumn{2}{c}{\textbf{Coverage}} & \multicolumn{2}{c}{\textbf{NIST16}} & \multicolumn{2}{c}{\textbf{OpenFor.}} & \multicolumn{2}{c}{\textbf{CocoGl.}} & \multicolumn{2}{c}{\textbf{DID}} & \multicolumn{2}{c}{\textbf{Overall}} \\
     \cmidrule(lr){3-4} \cmidrule(lr){5-6} \cmidrule(lr){7-8} \cmidrule(lr){9-10} \cmidrule(lr){11-12} \cmidrule(lr){13-14} \cmidrule(lr){15-16} \cmidrule(lr){17-18}
     \multicolumn{2}{c}{} & F1 & AUC & F1 & AUC  & F1 & AUC & F1 & AUC  & F1 & \multicolumn{1}{c}{AUC} & F1 & AUC & F1 & AUC & F1 & AUC    \\ 
     \midrule
     \parbox[t]{2mm}{\multirow{7}{*}{\rotatebox[origin=c]{90}{Feature Fusion}}} & \textbf{SPAN} \cite{hu2020span} & 21.4 & 83.3 & 72.5 & \textbf{98.1} & 42.5 & 86.5 &  0.4 & 56.2 &  0.2 & 52.8 & 20.8 & 75.0 & 7.5 & 61.5 & 23.6 & 73.3 \\
     & \textbf{IFOSN} \cite{wu2022robust}                      & 58.9 & 90.3 & 51.9 & 89.9 & 20.1 & 68.6 & 25.6 & 76.1 & 16.1 & 66.3 & 34.7 & 74.8 & 11.7 & 55.8 & 31.3 & 74.5 \\
     & \textbf{PSCC-Net} \cite{liu2022pscc}                   & 62.1 & 84.2 & 26.2 & 66.8 & 29.2 & 80.1 & 10.9 & 58.8 & 18.1 & 60.8 & \underline{48.2} & 83.3 & 44.6 & 74.7 & 34.2 & 72.7 \\
     & \textbf{MVSSNet++} \cite{dong2022mvss}                  & 58.5 & 81.9 & 63.1 & 86.9 & 46.0 & 85.9 & 24.5 & \underline{78.3} & 20.3 & 69.6 & 44.5 & 81.5 & 19.1 & 56.0 & 39.4 & 77.1 \\
     & \textbf{TruFor} \cite{guillaro2023trufor}               & \underline{71.7} & 95.7 & 67.6 & 95.9 & \underline{52.7} & \underline{89.4} & \underline{28.2} & 76.0 & 69.4 & \underline{88.8} & 44.3 & \underline{86.4} & 42.9 & 83.2 & 53.8 & \underline{87.9} \\
     & \textbf{CATNetv2} \cite{kwon2022learning}               & 70.2 & \underline{96.3} & \underline{83.4} & 95.2 & 35.0 & 75.7 & 16.0 & 62.4 & \underline{70.1} & 81.8 & 41.2 & 78.0 & \underline{75.2} & 95.2 & \underline{55.9} & 83.5 \\
     & \textbf{OMG-Fuser$_F$} (Ours)                               & \textbf{84.5} & \textbf{97.2} & \textbf{86.1} & \underline{97.2} & \textbf{63.1} & \textbf{91.4} & \textbf{34.2} & \textbf{79.1} & \textbf{72.2} & \textbf{92.1} & \textbf{53.9} & \textbf{88.3} & \textbf{77.0} & \textbf{95.4} & \textbf{67.3} & \textbf{91.5} \\
     \midrule
     \parbox[t]{2mm}{\multirow{4}{*}{\rotatebox[origin=c]{90}{Score Fus.}}} & \textbf{DST-Fusion} \cite{fontani2013framework}          & 75.3 & 94.8 & 85.4 & 93.5 & 48.8 & 79.2 & 14.7 & 56.9 & 20.4 & 50.7 & 38.6 & 80.9 & 24.9 & 72.8 & 44.0 & 75.6 \\
     & \textbf{AVG-Fusion} & 77.5 & \underline{97.3} & \underline{87.6} & \textbf{98.9} & \underline{52.1} & \underline{91.3} & 18.5 & \textbf{83.1} & 27.2 & \underline{90.8} & 42.8 & \underline{87.8} & 19.0 & \underline{90.5} & 46.4 & \underline{91.4} \\
     & \textbf{OW-Fusion} \cite{charitidis2021operation}        & \underline{78.8} & 97.0 & 85.8 & 96.0 & 47.7 & 88.5 & \underline{31.7} & 74.4 & \underline{70.7} & 87.4 & \underline{48.0} & 80.7 & \underline{53.7} & 90.2 & \underline{59.5} & 87.7 \\
     & \textbf{OMG-Fuser$_S$} (Ours)                                  & \textbf{85.1} & \textbf{98.1} & \textbf{92.9} & \underline{98.7} & \textbf{70.1} & \textbf{96.0} & \textbf{37.5} & \underline{82.1} & \textbf{74.1} & \textbf{94.1} & \textbf{56.2} & \textbf{89.4} & \textbf{76.6} & \textbf{96.1} & \textbf{70.4} & \textbf{93.5} \\
     \bottomrule
\end{tabular}
    }
    \caption{Comparison on image forgery localization. Pixel-level F1 and AUC scores are presented for each algorithm and dataset. The best value per column is highlighted in bold, and the second best is underlined.}
    \label{table:forgery_localization_results}
\end{table*}
\endgroup

\textbf{Stream Drop:} On given training sets, some signals perform better than others. Thus, if we naively combine all streams, the network will learn to over-attend almost exclusively on the best-performing ones in the training set, which can significantly limit the generalization performance of the model. To mitigate this issue, inspired by the idea of DropPath \cite{huang2016deep} proposed for dropping residual paths, we introduce the \emph{Stream Drop} (SD) mechanism for dropping entire streams of multi-stream networks during training to force the network to learn to capture useful information out of any input stream. To this end, each token in \cref{eq:streams_concat} is redefined based on the SD mechanism, which randomly drops a stream with probability $p_{drop}$. SD can be formulated as
\begin{equation}
    z_k = 
        \begin{cases}
            z_k / p_{drop}  &\quad P(1-p_{drop}) \\
            0 &\quad P(p_{drop}) \\
        \end{cases}.
\end{equation}

\begin{table*}
    \centering
    \scalebox{0.81}{
        \begin{tabular}{c l | cc cc cc cc cc | cc}
     \toprule
     \multicolumn{2}{c}{\multirow{2}{*}{\textbf{Approach}}}& \multicolumn{2}{c}{\textbf{CASIAv1+}} & \multicolumn{2}{c}{\textbf{Columbia}} & \multicolumn{2}{c}{\textbf{Coverage}} & \multicolumn{2}{c}{\textbf{NIST16}} & \multicolumn{2}{c}{\textbf{CocoGl.}} & \multicolumn{2}{c}{\textbf{Overall}} \\
     \cmidrule(lr){3-4} \cmidrule(lr){5-6} \cmidrule(lr){7-8} \cmidrule(lr){9-10} \cmidrule(lr){11-12} \cmidrule(lr){13-14}
     \multicolumn{2}{c}{} &   F1 &   AUC  &  F1  &  AUC  &  F1  &  AUC  &  F1  &  \multicolumn{1}{c}{AUC}  &  F1   & AUC & F1 & AUC \\
     \hline
     \parbox[t]{2mm}{\multirow{7}{*}{\rotatebox[origin=c]{90}{Feature Fusion}}} & \textbf{PSCC-Net} \cite{liu2022pscc}                   & 54.0 & 86.9 & 26.6 & 80.8 & 29.7 & 65.7 & 63.6 & 62.0 & 52.9 & 77.8 & 45.4 & 74.6 \\
     & \textbf{TruFor} \cite{guillaro2023trufor}              & 80.1 & 91.7 & \underline{98.3} & 99.6 & 56.2 & 77.0 & 28.7 & 62.7 & 48.6 & 75.2 & 62.4 & \underline{81.2} \\
     & \textbf{SPAN} \cite{hu2020span}     & 61.8 & 74.5 & 98.4 & \textbf{99.9} & \underline{75.8} & \underline{82.6} & 16.3 & 56.7 & \underline{71.7} & \underline{78.0} & 64.8 & 78.3 \\ 
     & \textbf{IFOSN} \cite{wu2022robust}  & 69.7 & 73.9 & 67.3 & 88.2 & 65.7 & 55.7 & 67.2 & 66.4 & 62.2 & 61.1 & 66.4 & 69.0 \\
     & \textbf{MVSSNet++} \cite{dong2022mvss}                & 72.2 & 85.5 & 78.1 & 98.3 & 66.1 & 71.3 & 65.8 & 58.9 & 67.6 & 67.9 & 70.0 & 76.4 \\   
     & \textbf{CATNetv2} \cite{kwon2022learning}              & \underline{86.1} & \underline{94.4} & 83.1 & 95.2 & 64.0 & 67.9 & \underline{69.4} & \underline{75.0} & 64.8 & 66.6 & \underline{73.5} & 79.8 \\ 
     & \textbf{OMG-Fuser$_F$} (Ours)                                      & \textbf{90.7} & \textbf{96.5} & \textbf{99.2} & \textbf{99.9} & \textbf{78.8} & \textbf{83.5} & \textbf{71.4} & \textbf{78.2} & \textbf{72.0} & \textbf{81.7} & \textbf{82.4} & \textbf{88.0} \\ 
     \midrule
      \parbox[t]{2mm}{\multirow{4}{*}{\rotatebox[origin=c]{90}{Score Fus.}}} & \textbf{DST-Fusion} \cite{fontani2013framework}        & 80.3 & \underline{93.83} & 92.3 & 97.9 & 69.0 & 78.4 & 44.4 & 75.1 & 66.0 & 78.5 & 70.4 & 84.7 \\
      & \textbf{OW-Fusion} \cite{charitidis2021operation}      & 85.3 & 93.3 & 80.1 & 97.7 & 66.1 & 72.4 & \underline{69.8} & 75.5 & \underline{69.3} & 72.2 & 74.1 & 82.2 \\
     & \textbf{AVG-Fusion} & \underline{86.1} & 93.0 & \underline{94.5} & \underline{99.7} & \underline{73.3} & \underline{80.7} & 69.0 & \underline{76.7} & 69.2 & \underline{78.2} & \underline{78.4} & \underline{85.7} \\
     & \textbf{OMG-Fuser$_S$} (Ours)   & \textbf{91.0} & \textbf{98.0} & \textbf{99.4} & \textbf{99.9} & \textbf{81.0} & \textbf{84.5} & \textbf{71.9} & \textbf{80.2} & \textbf{72.6} & \textbf{82.7} & \textbf{83.2} & \textbf{89.5} \\ 
     \bottomrule
\end{tabular}
    }
    \caption{Comparison on image forgery detection. Image-level F1 and AUC scores are presented for each algorithm and dataset. The best value per column is highlighted in bold, and the second best is underlined.}
    \label{table:forgery_detection_results}
\end{table*}
\section{Experiments}\label{sec:experiments}

\subsection{Experimental setup}
OMG-Fuser provides a network for combining both the score-level outputs of multiple image forensics algorithms as well as multiple low-level image forensic signals. Thus, we experiment with two different variants of our architecture: (i) one that incorporates five recent forgery localization algorithms \cite{hu2020span, wu2022robust, dong2022mvss, kwon2022learning, guillaro2023trufor} (OMG-Fuser$_S$), and (ii) one that directly fuses two recently proposed learnable forensic cues, namely the DCT stream  \cite{kwon2022learning} and the Noiseprint++\cite{guillaro2023trufor} at the feature-level (OMG-Fuser$_F$). We train both variants on 40k forged samples and 32k authentic samples compiled from the datasets provided in \cite{kwon2021cat, mahfoudi2019defacto, dong2013casia}, while we evaluate them on CASIAv1+ \cite{dong2022mvss}, Columbia \cite{hsu2006detecting}, Coverage \cite{wen2016coverage}, NIST16 \cite{guan2019mfc}, OpenForensics \cite{le2021openforensics}, CocoGlide \cite{guillaro2023trufor} and DID \cite{wu2021iid} datasets. We employ six recent competitor methods with publicly available implementations for the feature fusion category \cite{hu2020span, liu2022pscc, wu2022robust, dong2022mvss, kwon2022learning, guillaro2023trufor}. Regarding score-level fusion, we reimplement two popular methods that support the fusion of multiple signals \cite{charitidis2021operation, fontani2013framework}, along with a baseline average fusion approach, using the same input signals as OMG-Fuser$_S$. We use the SAM~\cite{kirillov2023segment} pretrained model for instance segmentation. DINOv2~\cite{oquab2023dinov2} model is used as the vision backbone, which is fine-tuned on our OMG-Fuser$_F$ variant, but kept frozen on OMG-Fuser$_S$. Moreover, we employ the F1-score with a threshold of $0.5$ as the main indicator of performance when deployed in the wild and the AUC as an auxiliary threshold-agnostic metric. More information regarding implementation details, datasets, and algorithms, along with localization results in terms of the F1 metric using the best threshold per image, is provided in the supplementary material. \looseness=-1

\subsection{Comparison with the state-of-the-art}
\textbf{Image forgery localization} evaluation results are presented in \cref{table:forgery_localization_results}. The reported F1 and AUC metrics are computed on the pixel-level. We see that our implementations outperform all state-of-the-art methods in score- and feature-level fusion. Moreover, in the case of score-level fusion, we show that previous approaches, either based on some statistical frameworks or learning approaches without semantic information, perform poorly on the recent deep learning-based forensic signals, even worse than averaging the signals. This effectively shows that our network is capable of effectively fusing multiple forensic signals by exploiting the object-level information of the image, and at the same time, it is capable of handling low-level forensic traces. The high performance in feature-level fusion enables future works on new forensic signals to use our network to decrease the burden of developing purpose-built fusion architectures.

\begin{figure*}
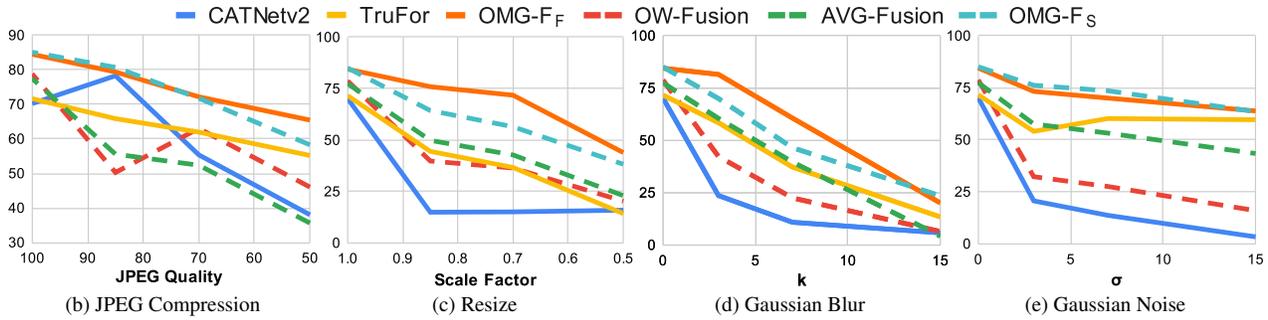

    \centering
    \subfloat{\includesvg[width=0.7\textwidth]{figures/source/robustness_legend.svg}} \\
    \subfloat[JPEG Compression]{\includesvg[width=0.235\textwidth]{figures/source/jpeg_compression.svg}}
    \subfloat[Resize]{\includesvg[width=0.24\textwidth]{figures/source/resize.svg}}
    \subfloat[Gaussian Blur]{\includesvg[width=0.24\textwidth]{figures/source/gauss_blur.svg}}
    \subfloat[Gaussian Noise]{\includesvg[width=0.24\textwidth]{figures/source/gauss_noise.svg}}
    \caption{Robustness evaluation on common perturbations. The pixel-level F1 is reported. Straight lines denote the feature-level approaches, and dashed lines the score-level approaches. The top approaches of each category are shown for readability.}
    \label{fig:robustness}
\end{figure*}

\begin{table}
    \centering
    \scalebox{0.85}{
    \begin{tabular}{c l | cc | cc }
     \toprule
     \multicolumn{2}{c}{\multirow{2}{*}{\textbf{Ablation Study}}} & \multicolumn{2}{c}{\textbf{Loc.}} & \multicolumn{2}{c}{\textbf{Det.}} \\
     \cmidrule(lr){3-4} \cmidrule(lr){5-6}
     \multicolumn{2}{c}{} & F1 & \multicolumn{1}{c}{AUC} & F1 & AUC \\ \midrule \midrule
     
     \multicolumn{2}{l|}{\textbf{OMG-Fuser$_S$}}           & 70.4 & 93.5 & 83.2 & 89.5 \\
      \midrule
      \parbox[t]{2mm}{\multirow{6}{*}{\rotatebox[origin=c]{90}{Signals}}} & w/o SPAN   & 69.2 & 91.5 & 82.3 & 89.1 \\
     & w/o IFOSN            & 68.5 & 91.1 & 81.5 & 88.9 \\
     & w/o MVSSNet++        & 67.6 & 90.2 & 81.0 & 88.3 \\
     & w/o CATNetv2         & 63.6 & 89.7 & 78.1 & 87.6 \\
     & w/o TruFor           & 63.5 & 89.1 & 78.2 & 85.8 \\
     & w/o RGB              & 53.6 & 87.0 & 76.8 & 84.7 \\
     \midrule
     \parbox[t]{2mm}{\multirow{4}{*}{\rotatebox[origin=c]{90}{Compon.}}} & w/o OGA    & 61.1 & 88.7 & 79.7 & 85.8 \\
     & w/o TFT              & 63.5 & 88.4 & 78.1 & 86.3 \\
     & w/o LDT              & 60.0 & 89.3 & 79.6 & 85.7 \\
     & w/o Stream Drop      & 66.5 & 90.5 & 81.1 & 87.3 \\
     \midrule \midrule
     \multicolumn{2}{l|}{\textbf{OMG-Fuser$_F$}} & 67.3 & 91.5 & 82.4 & 88.0 \\
     \midrule
     \parbox[t]{2mm}{\multirow{3}{*}{\rotatebox[origin=c]{90}{Signals}}} & w/o Noiseprint++ & 60.7 & 85.2 & 72.9 & 83.4 \\
     & w/o DCT              & 62.0 & 85.3 & 73.4 & 83.7 \\
     & w/o RGB              & 64.3 & 82.8 & 77.6 & 84.7 \\ 
     \bottomrule
\end{tabular}
    }
    \caption{Ablation Study. The average pixel-level F1 and AUC scores are reported across all evaluation datasets.}
    \label{table:ablation_studies}
\end{table}

\begin{figure}
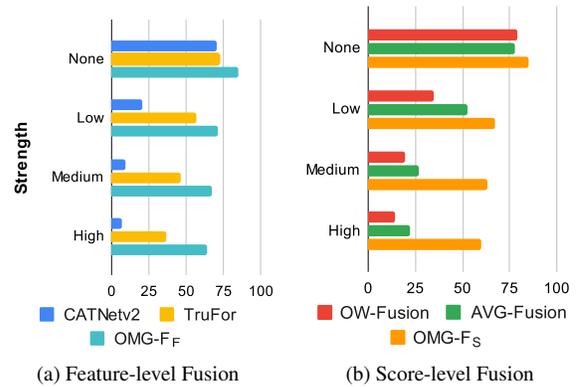

    \centering
    \subfloat[Feature-level Fusion]{\includesvg[width=0.45\columnwidth]{figures/source/jpeg_artifacts_removal_ff.svg}} \hspace{5pt}
    \subfloat[Score-level Fusion]{\includesvg[width=0.45\columnwidth]{figures/source/jpeg_artifacts_removal_sf.svg}}
    \caption{Robustness against neural filters for removing JPEG artifacts. The pixel-level F1 is reported.}
    \label{fig:robustness_jpeg_artifacts}
\end{figure}

\textbf{Image forgery detection} evaluation results are presented in \cref{table:forgery_detection_results}. Following \cite{guillaro2023trufor}, for methods that do not output an image-level forgery detection score, we compute it as the max of the corresponding forgery localization mask. Also, we omit datasets that contain only forged samples. Both OMG-Fuser implementations outperform all state-of-the-art methods by a significant margin.

\subsection{Ablation Study}
In order to better understand the contribution of each of the major components of our architecture, we performed several ablation studies in two directions. First, we evaluated the contribution of each of the fused forensic signals by training our network from scratch and removing the inputs of the network one at a time. Similarly, we proceeded by evaluating the contribution of each of the key components of our architecture by: replacing the OGA with the standard unmasked self-attention \cite{vaswani2017attention}, replacing the TFT with an average pooling layer, removing the LDT and Stream Drop components. To better understand the impact of each component on signal fusion, we performed the later study on the OMG-Fuser$_S$, which fuses the output of several forensic algorithms. The results of the ablation study are presented in \cref{table:ablation_studies}. Removing a single input signal affects the performance of the network adversely. Thus, the network is capable of learning to exploit even minor additional information introduced by new input streams. Moreover, we see that removing any key component of the architecture considerably impacts performance, highlighting that they are of crucial importance for effective signal fusion.

\subsection{Robustness against Perturbations and Filters}

To assess the robustness of our architecture against common online perturbations, we performed another set of experiments, where we applied different levels of JPEG compression, resizing, Gaussian noise, and Gaussian blur to the input images. We used the challenging CASIAv1+ dataset, as it contains already compressed samples with many different depicted objects. The results of this study for the task of image forgery localization are presented in \cref{fig:robustness}, where the pixel-level F1 is reported. We see that the two variants of our architecture consistently outperform the state-of-the-art methods, both in feature-level and score-level fusion. Moreover, we evaluate our implementations against the recently added neural filter in Adobe Photoshop \cite{psjpegartifacts2023} for removing JPEG artifacts and present the results in \cref{fig:robustness_jpeg_artifacts}. Our method demonstrates its robustness even in this challenging forgery attack, outperforming all competing approaches. \looseness=-1

\setlength{\textfloatsep}{1pt}{
\begin{figure*}
  \centering
  \includesvg[width=0.90\textwidth]{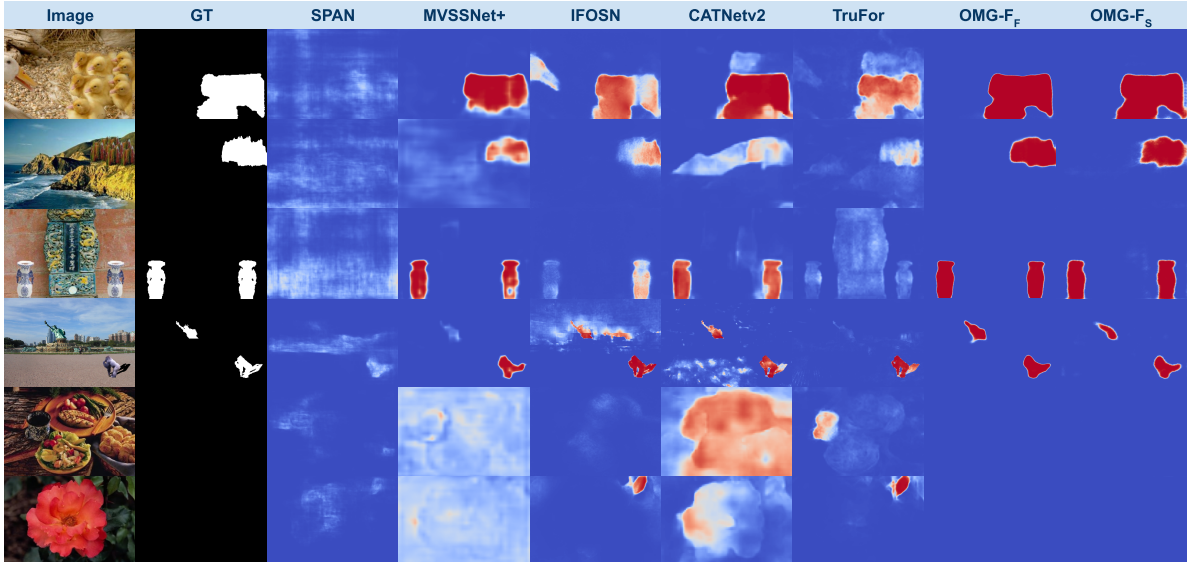}
  \caption{Qualitative evaluation results. From left to right, the image in question, the ground truth mask, the outputs of five recent methods, and the outputs of the two variants of our architecture are displayed.
  \vspace{3pt}}
  \label{fig:qualitative_eval}
\end{figure*}
}
\vspace{-3pt}

\subsection{Stream Expansion}
The design of the TFT allows the expansion of an already trained model with additional streams. In \cref{table:stream_expansion}, we report the overall performance across all the evaluation datasets when expanding our architecture in three different ways. Starting from a variant of OMG-Fuser$_S$ trained without the TruFor stream, we expand it i) by training only the additional stream and keeping the rest of the network frozen, ii) by fine-tuning the network for 15\% of the initial training epochs, and iii) by training the network from scratch. We see that in the first two cases, the network achieves a performance very close to third by requiring only a portion of the computational resources. This suggests that our model has learned to fuse various forensic signals effectively; therefore, it can be expanded with new signals without training from scratch. \looseness=-1

\begin{table}
    \centering
    \scalebox{0.85}{
    \begin{tabular}{l | cc | cc}
     \toprule
     \multicolumn{1}{c}{\multirow{2}{*}{\textbf{Expansion}}} & \multicolumn{2}{c}{\textbf{Loc.}} & \multicolumn{2}{c}{\textbf{Det.}} \\
     \cmidrule(lr){2-3} \cmidrule(lr){4-5}
     \multicolumn{1}{c}{} & F1 & \multicolumn{1}{c}{AUC} & F1 & AUC \\
     \midrule
     4-stream model          & 63.5 & 89.1 & 78.2 & 85.8 \\ \midrule
     stream-only train       & 68.8 & 92.1 & 82.4 & 88.6 \\
     full fine-tuning        & 69.0 & 92.7 & 82.7 & 89.1 \\ \midrule
     training from scratch   & 70.4 & 93.5 & 83.2 & 89.5 \\
     \bottomrule
\end{tabular}
    }
    \caption{Evaluation of expanding the network with a new stream. The average scores are reported across all the evaluation datasets.
    \vspace{3pt}}
    \label{table:stream_expansion}
\end{table}

\subsection{Qualitative Evaluation}

In \cref{fig:qualitative_eval}, we present a qualitative evaluation of the two variants of our architecture on samples containing one, two, or no forged regions. In all the cases, both our variants lead to more robust results than all competing methods, clearly identifying the forged regions of the image and reducing false positives. The fusion process has a minimal effect on computation time, adding less than 100ms to the total process executed on commodity hardware. More details about timings are provided in the supplementary material. \looseness=-1
\section{Conclusions}\label{sec:conclusions}

In this paper, we introduced a novel transformer-based network architecture for fusing an arbitrary number of image forensic signals based on the object-level information of the image. We demonstrated that both the effective fusion of signals and object-level information is essential for robust forensic analysis. Also, the modularity of our architecture was shown to be effective for score-level fusion of five recent image forgery localization algorithms and feature-level fusion of two recently proposed learnable forensic cues, outperforming all state-of-the-art methods on the tasks of image forgery detection and localization, while being robust to several traditional and novel forensic attacks. Finally, our network can facilitate future work as its expansion with new signals does not necessitate training from scratch.

\small\medskip\noindent\textbf{Acknowledgments:}
This work was supported by the Horizon Europe vera.ai project (grant no.101070093), Junior Star GACR (grant no. GM 21-28830M), and EuroHPC for providing access to MeluXina supercomputer (grant no. EHPC-DEV-2023D03-008).

\clearpage
{
    \small
    \bibliographystyle{ieeenat_fullname}
    \bibliography{main}
}

\twocolumn[\centerline{\parbox[c][3cm][c]{12cm}{
            \centering
            \fontsize{16}{16}
            \selectfont
            {Supplementary Material}}}]

\appendix
\section{Experimental Setup}

\subsection{Image Forensics Algorithms} \label{sec:algorithms}

In our evaluation, we consider only methods with publicly available code in order to fairly evaluate them under the same setup. We employ as inputs to our score-level fusion network, OMG-Fuser$_S$, the SPAN \cite{hu2020span}, MVSS-Net++ \cite{dong2022mvss}, CATNetv2 \cite{kwon2022learning}, TruFor \cite{guillaro2023trufor} and IFOSN \cite{wu2022robust} algorithms, due to their competitive performance and their complementarity with respect to the type of artifacts they detect, i.e., artifacts in the RGB domain, edge artifacts, compression artifacts, noise artifacts, and robustness against online sharing operations, respectively. Moreover, we utilize the Noiseprint++ \cite{guillaro2023trufor} and the DCT-domain stream~\cite{kwon2022learning} as inputs to our feature-level fusion network, OMG-Fuser$_F$, two learnable forensic signals that captures noise-related and compression-related artifacts, respectively. For all the aforementioned methods, we utilize the official code implementations provided by the original authors.

\subsection{Datasets} \label{sec:datasets}

To train our OMG-Fuser models, we utilize data from three publicly available datasets. We randomly sample 25k forged and 25k authentic images from the synthetic dataset used by CAT-Net~\cite{kwon2021cat} due to the big variability of its samples regarding compression qualities and depicted topics. We enrich this set with another 10k inpainted images sampled from the DEFACTO~\cite{mahfoudi2019defacto} dataset. Also, we include all images from the CASIAv2 \cite{dong2013casia} dataset into our training set to compensate for low-resolution and low-quality images. We utilize 90\% of these data for the actual training of the model and the rest 10\% for validation purposes. Despite some datasets consisting of more samples, we observed in our experiments that a further increase in the amount of training data did not yield any significant performance increase. Thus, our architecture requires significantly less training data than the previous state-of-the-art ones \cite{guillaro2023trufor, kwon2022learning}. Instead, as highlighted by \cite{guillaro2023trufor}, introducing more variability into the training data was more beneficial. For evaluation, following \cite{hu2020span, liu2022pscc, guillaro2023trufor}, we have selected five popular datasets, namely the CASIAv1+ \cite{dong2022mvss}, COLUMBIA \cite{hsu2006detecting}, COVERAGE \cite{wen2016coverage}, NIST16 \cite{guan2019mfc}, OpenForensics \cite{le2021openforensics} datasets, including common cases of image forgery. In contrast to previous works~\cite{hu2020span, liu2022pscc}, we benchmark our methods on the entire evaluation dataset without making any further assumptions about the type of forgery or utilizing subsets of them. Furthermore, to take into account deep-learning based inpainting operations, we further employed the recently introduced CocoGlide \cite{guillaro2023trufor} dataset and the deep-learning based inpaintings of the DiverseInpaintingDataset (DID) \cite{wu2021iid}.  %, in order to better resemble the in-the-wild performance
A summary of all the datasets used in our research is presented in \cref{table:datasets}.

\begin{table}[]
    \centering
    \scalebox{0.85}{
    \begin{tabular}{c l c c c}
     \toprule
     \multicolumn{2}{c}{\textbf{Dataset}}         & \textbf{Forged} & \textbf{Authentic} & \textbf{Types} \\
     \midrule
     \parbox[t]{2mm}{\multirow{3}{*}{\rotatebox[origin=c]{90}{Train}}} & \textbf{Tampered-50k} \cite{kwon2021cat}          & 25k & 25k & SP, CM      \\
     & \textbf{DEFACTO-INP} \cite{mahfoudi2019defacto} & 10k & -   & INP         \\
     & \textbf{CASIAv2} \cite{dong2013casia}           & 5k  & 7k  & SP, CM      \\
     \midrule
      \parbox[t]{2mm}{\multirow{7}{*}{\rotatebox[origin=c]{90}{Test}}} & \textbf{CASIAv1+} \cite{dong2022mvss}             & 920 & 800 & SP, CM      \\
     & \textbf{Columbia} \cite{hsu2006detecting}         & 180 & 183 & SP          \\
     & \textbf{COVERAGE} \cite{wen2016coverage}          & 100 & 100 & CM          \\
     & \textbf{NIST16}  \cite{guan2019mfc}               & 564 & 560 & SP, CM, INP \\
     & \textbf{OpenForensics} \cite{le2021openforensics} & 19k & -   & SP          \\
     & \textbf{CocoGlide} \cite{guillaro2023trufor}      & 512 & 512 & INP         \\
     & \textbf{DID} \cite{wu2021iid}                     & 6k  & -   & INP         \\
     \bottomrule
\end{tabular}

    }
    \caption{Number of samples and types of forgery included in the train and test datasets. SP stands for splicing, CM for copy-move and INP for inpainting.}
    \label{table:datasets}
\end{table}

\subsection{Implementation Details} \label{sec:impl_details}

We implement and train all of our models using PyTorch ~\cite{paszke2019pytorch}. Following~\cite{wilson2017marginal, naganuma2022empirical}, we train our models for 100 epochs using the SGD optimizer with momentum~\cite{liu2020improved} set to 0.9. We initialize the learning rate to $10^{-3}$, with 5 epochs of linear warm-up and a cosine decay until $10^{-6}$. We empirically tune the weights of the optimization criterion and set them to $a=0.3$, $b=0.45$, $c=0.25$. To acquire the instance segmentation masks, we utilize the Segment Anything Model (SAM) \cite{kirillov2023segment}, a zero-shot model that is not limited to a fixed set of object classes. Furthermore, for the RGB stream, we employ the DINOv2 model \cite{oquab2023dinov2}, trained in an unsupervised manner on a large curated dataset and capable of extracting rich features suitable for a large number of downstream tasks. We utilize its ViT-S/14 variant. During the training of score-level fusion models, DINOv2 remains frozen and the resolution of inputs to all streams is $224 \times 224$. For feature-level fusion models, we increase the input resolution of all streams to $448 \times 448$ and fine-tune the DINOv2 backbone in order to capture low-level cues in finer detail. Following~\cite{kumar2022fine, lee2019would}, we freeze the patch-embedding layer during fine-tuning. For the computation of the input signals, the image is provided in its original resolution to all the respective algorithms. 

The number of layers of each stage is set to $B_1 = B_2 = B_3 = 6$. Regarding the localization head, it consists of five upsampling layers, each including a transposed convolution \cite{long2015fully}, a ReLU \cite{krizhevsky2012imagenet} activation, and a Batch Normalization \cite{ioffe2015batch} layer, with a sigmoid activation at the end of the network. For the detection head, we employ a four-block transformer~\cite{dosovitskiy2020image} with a classification token $z^{cls} \in \mathbb{R}^D$ that is used for forgery detection. After propagating through the network, the refined token passes through a single fully-connected layer with a sigmoid activation to generate the final image-level forgery detection score. 

The training data are augmented using resizing, cropping, flipping, and rotation operations. Training is performed on a single HPC cluster node equipped with four Nvidia A100 40GB GPUs, with an effective batch size of 160 images for score-level fusion models and 40 for feature-level fusion models. The training requires about 30 hours for the score-level and 60 hours for the feature-level fusion models. Moreover, the stream expansion experiments are performed on a single A100 to better represent a constrained environment. Finally, all the evaluation experiments are being conducted on a single Nvidia RTX3090 GPU.

\begin{table*}
    \centering
    \scalebox{0.99}{
    \begin{tabular}{ c l | c c c c c c c | c }
    \toprule
    \multicolumn{2}{c}{\textbf{Approach}} & \textbf{CASIAv1+} & \textbf{Columbia} & \textbf{Coverage} & \textbf{NIST16} & \multicolumn{1}{c}{\textbf{OpenFor.}} & \textbf{CocoGl.} & \textbf{DID} & \textbf{Overall} \\
     \midrule
     \parbox[t]{2mm}{\multirow{7}{*}{\rotatebox[origin=c]{90}{Feature Fusion}}} & \textbf{PSCC-Net} \cite{liu2022pscc}  & 83.4 & 86.7 & 69.3 & 50.2 & 51.3 & 84.1 & 58.5 & 70.7 \\
     & \textbf{SPAN} \cite{hu2020span} & 64.2 & 88.6 & 78.6 & 56.6 & 39.0 & 82.7 & 45.9 & 71.3 \\
     % & \textbf{Mantranet} \cite{wu2019mantra}          & & & & & & \\
     & \textbf{IFOSN} \cite{wu2022robust}                & 87.4 & 86.9 & 63.4 & 71.1 & 49.3 & 79.9 & 37.8 & 73.4 \\
     & \textbf{MVSSNet++} \cite{dong2022mvss}            & 80.0 & 81.6 & 80.0 & 73.1 & 48.1 & 83.3 & 41.5 & 75.1 \\
     & \textbf{CATNetv2} \cite{kwon2022learning}         & 87.6 & \underline{91.7} & 79.0 & 68.9 & 66.3 & 80.3 & \textbf{80.0} & 77.2 \\
     & \textbf{TruFor} \cite{guillaro2023trufor}         & \underline{89.6} & 90.5 & \underline{83.9} & \underline{74.5} & \underline{71.2} & \underline{85.6} & 66.5 & \underline{77.8} \\
     & \textbf{OMG-Fuser$_F$} (Ours)                     & \textbf{92.0} & \textbf{94.6} & \textbf{88.3 }& \textbf{82.1} & \textbf{82.0} & \textbf{87.7} & \underline{78.9} & \textbf{80.1} \\
     \midrule
     \parbox[t]{2mm}{\multirow{4}{*}{\rotatebox[origin=c]{90}{Score Fus.}}} & \textbf{DST-Fusion} \cite{fontani2013framework} & 89.3 & 91.8 & 76.6 & 56.0 & 33.3 & 86.5 & 64.8 & 77.4 \\
     & \textbf{OW-Fusion} \cite{charitidis2021operation} & 89.1 & \underline{94.7} & 82.0 & 72.5 & 66.9 & 84.2 & \underline{78.2} & 81.9 \\
     & \textbf{AVG-Fusion}                               & \underline{90.9} & 94.4 & \underline{87.8} & \underline{80.4} & \underline{70.9} & \textbf{87.8} & 77.5 & \underline{84.3} \\
     & \textbf{OMG-Fuser$_S$} (Ours)                     & \textbf{92.2} & \textbf{96.6} & \textbf{89.3} & \textbf{83.7} & \textbf{85.2} & \underline{86.2} & \textbf{86.1} & \textbf{86.8} \\
     \bottomrule
\end{tabular}
    }
    \caption{Comparison on image forgery localization. Pixel-level F1 scores, calibrated with the best threshold per image, are presented for each algorithm and dataset. The best value per column is highlighted in bold, and the second best is underlined.}
    \label{table:f1best_forgery_localization}
\end{table*}

For comparison with other score-level fusion approaches, we employ the OW-Fusion \cite{charitidis2021operation}, a deep learning-based fusion approach, and implement it with the same input signals used on OMG-Fuser$_S$. Furthermore, in order to take into account the previous statistical fusion frameworks, we reimplement a DST-based fusion framework \cite{fontani2013framework, phan2022comparative} in Python, using again the same inputs with our score-level fusion implementation. Finally, we use the average of all input signals as a baseline approach.

\section{Additional Experiments}

\textbf{Image Forgery Localization on best threshold:} Following \cite{guillaro2023trufor}, we conducted additional experiments on image forgery localization, computing the F1 metric for the best threshold per image as an indicator of the performance of the method when properly calibrated. The results are presented in \cref{table:f1best_forgery_localization}. Similar to the results in the main paper, both our methods outperform the competing approaches from the state-of-the-art in both feature- and score-level fusion with a clear margin.

\begin{table}
    \centering
    \scalebox{0.99}{
    \begin{tabular}{c l | cc | cc }
     \toprule
     \multicolumn{2}{c}{\multirow{2}{*}{\textbf{Seg. Model}}} & \multicolumn{2}{c}{\textbf{Loc.}} & \multicolumn{2}{c}{\textbf{Det.}} \\
     \cmidrule(lr){3-4} \cmidrule(lr){5-6}
     \multicolumn{2}{c}{} & F1 & \multicolumn{1}{c}{AUC} & F1 & AUC \\     
      \midrule
      \parbox[t]{2mm}{\multirow{4}{*}{\rotatebox[origin=c]{90}{Feat. Fus.}}} & EVA (LVIS) & 66.1 & 90.5 & 80.7 & 87.5 \\
     & EVA (COCO)    & 66.5 & 90.7 & 80.3 & 87.9 \\
     & EVA (COCO+LVIS)       & 67.0 & 91.3 & 80.8 & 88.0 \\
     & SAM (SA-1B)           & 67.3 & 91.5 & 82.4 & 88.0 \\
     \midrule
     \parbox[t]{2mm}{\multirow{4}{*}{\rotatebox[origin=c]{90}{Score Fus.}}} & EVA (LVIS) & 68.4 & 92.6 & 81.9 & 87.9 \\
     & EVA (COCO+LVIS)       & 68.8 & 93.0 & 82.4 & 88.6 \\
     & EVA (COCO)            & 69.1 & 93.0 & 81.8 & 88.4 \\
     & SAM (SA-1B)           & 70.4 & 93.5 & 83.2 & 89.5\\
     \bottomrule
\end{tabular}
    }
    \caption{Comparison with different instance segmentation models. The average pixel-level F1 and AUC scores are reported across all evaluation datasets. The training dataset of each instance segmentation model is reported in parentheses.}
    \label{table:segmentation_models}
\end{table}

\begin{figure*}
  \centering
  \includesvg[width=0.99\textwidth]{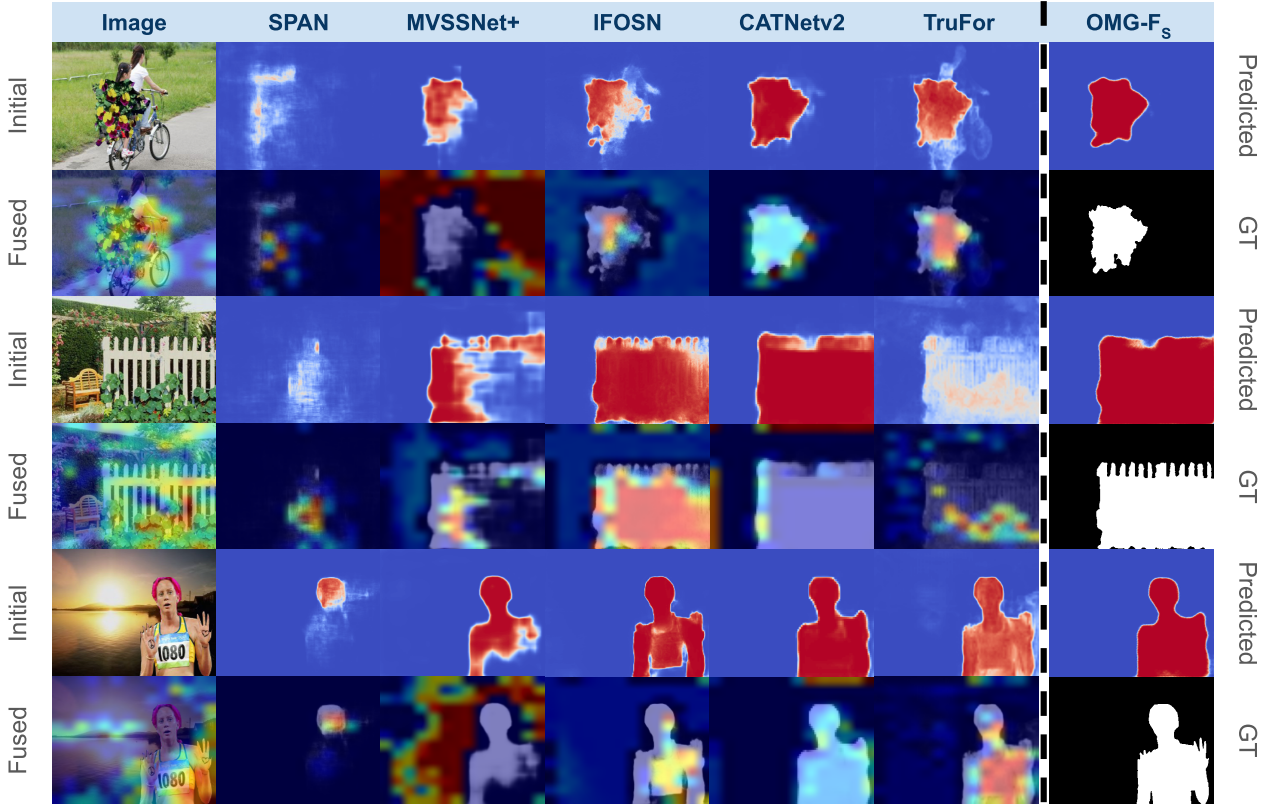}
  \caption{Explainability analysis. The top ``initial'' row of each sample presents the signals fused by OMG-F$_S$, while the bottom ``fused'' row presents the Grad-CAM overlay on top of them. Red regions in the overlay maps denote the regions of the signals with the greatest impact on the fusion process. The most-right column depicts the predicted output of our network on top and the ground-truth mask on the bottom row. \looseness=-1}
  \label{fig:explainability}
\end{figure*}

\textbf{Instance Segmentation Models:} To better evaluate the modularity of our architecture, starting from the trained models of our score- and feature-level fusion implementations, we replaced the instance segmentation masks generated by the SAM \cite{kirillov2023segment} with the ones computed by the EVA \cite{fang2023eva}. In particular, we considered three different types of instance segmentation masks: i) from an EVA model trained on COCO \cite{lin2014microsoft}, ii) from an EVA model trained on LVIS \cite{gupta2019lvis} and iii) from aggregating the segmentation masks of both of the aforementioned models. The results of these experiments are presented in \cref{table:segmentation_models}. They highlight that replacing the instance segmentation model used during training has only a minimal impact on performance. This allows the combination of our model with class-specific segmentation models that better fit the needs of the downstream application. \looseness=-1

\textbf{Computation Time:} In \cref{table:computation_times}, we present the computation time required for extracting the input signals for both our score- and feature-level fusion implementations. The experiments have been conducted on the NIST16 dataset due to its great variability in the sizes of the included images. Our proposed fusion networks impose a minimal overhead on the overall computation time compared to the computation time required for generating the fused signals.
 
\begin{table}
    \centering
    \scalebox{0.99}{
    \begin{tabular}{ c l | l c }
    \toprule
     \multicolumn{2}{c}{\textbf{Signal}} & \textbf{Time} \\
     \midrule
     \parbox[t]{2mm}{\multirow{4}{*}{\rotatebox[origin=c]{90}{Feat. Fus.}}} & \textbf{DCT} \cite{guillaro2023trufor} & 75 ms \\
     & \textbf{Noiseprint++} \cite{guillaro2023trufor} & 115 ms \\
     & \textbf{SegmentAnything} \cite{kirillov2023segment} & 1.4 s \\
     & \textbf{OMG-Fuser$_F$} (Ours)                     &  32.3 ms\\
     \midrule
     \parbox[t]{2mm}{\multirow{7}{*}{\rotatebox[origin=c]{90}{Score Fusion}}} & \textbf{SPAN} \cite{hu2020span} & 1.34s \\
     & \textbf{IFOSN} \cite{wu2022robust}                & 6.85 s \\
     & \textbf{MVSSNet++} \cite{dong2022mvss}            & 221 ms \\
     & \textbf{CATNetv2} \cite{kwon2022learning}         & 1.04 s  \\
     & \textbf{TruFor} \cite{guillaro2023trufor}         & 1.18 s \\
     & \textbf{SegmentAnything} \cite{kirillov2023segment} & 1.4 s \\
     & \textbf{OMG-Fuser$_S$} (Ours)                     & 40.4 ms \\
     \bottomrule
\end{tabular}
    }
    \caption{Computation time for the fused signals and our proposed network. The input signals utilized in the proposed feature- and score-level fusion implementations are considered. }
    \label{table:computation_times}
\end{table}

\begin{figure*}
  \centering
  \includesvg[width=0.99\textwidth]{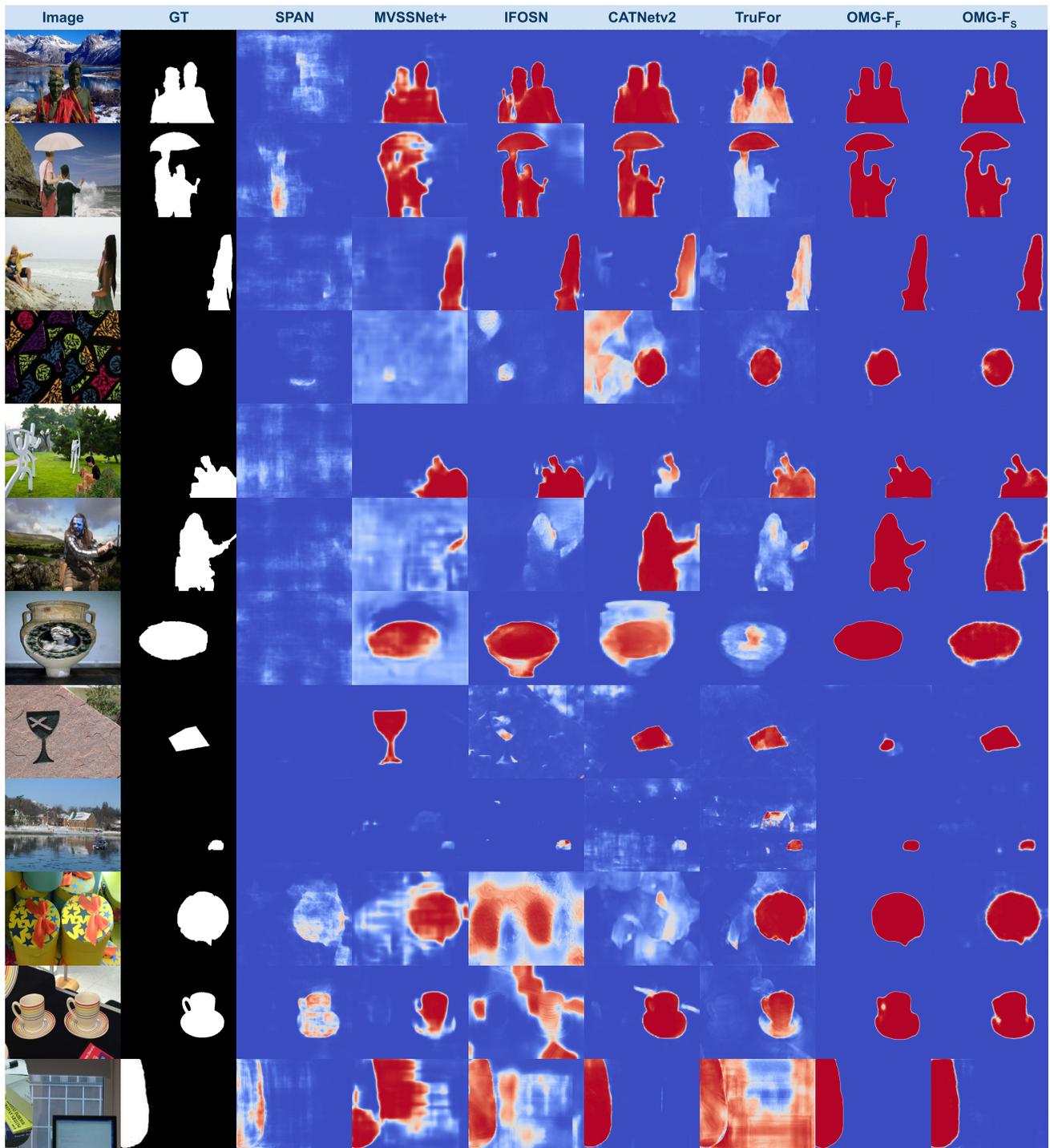}
  \caption{Additional qualitative evaluation results on forged images. From left to right, the image in question, the ground truth mask, the outputs of five recent methods, and the outputs of the two variants of our architecture are displayed.}
  \label{fig:extended_qual_eval_forged}
\end{figure*}

\begin{figure*}
  \centering
  \includesvg[width=0.99\textwidth]{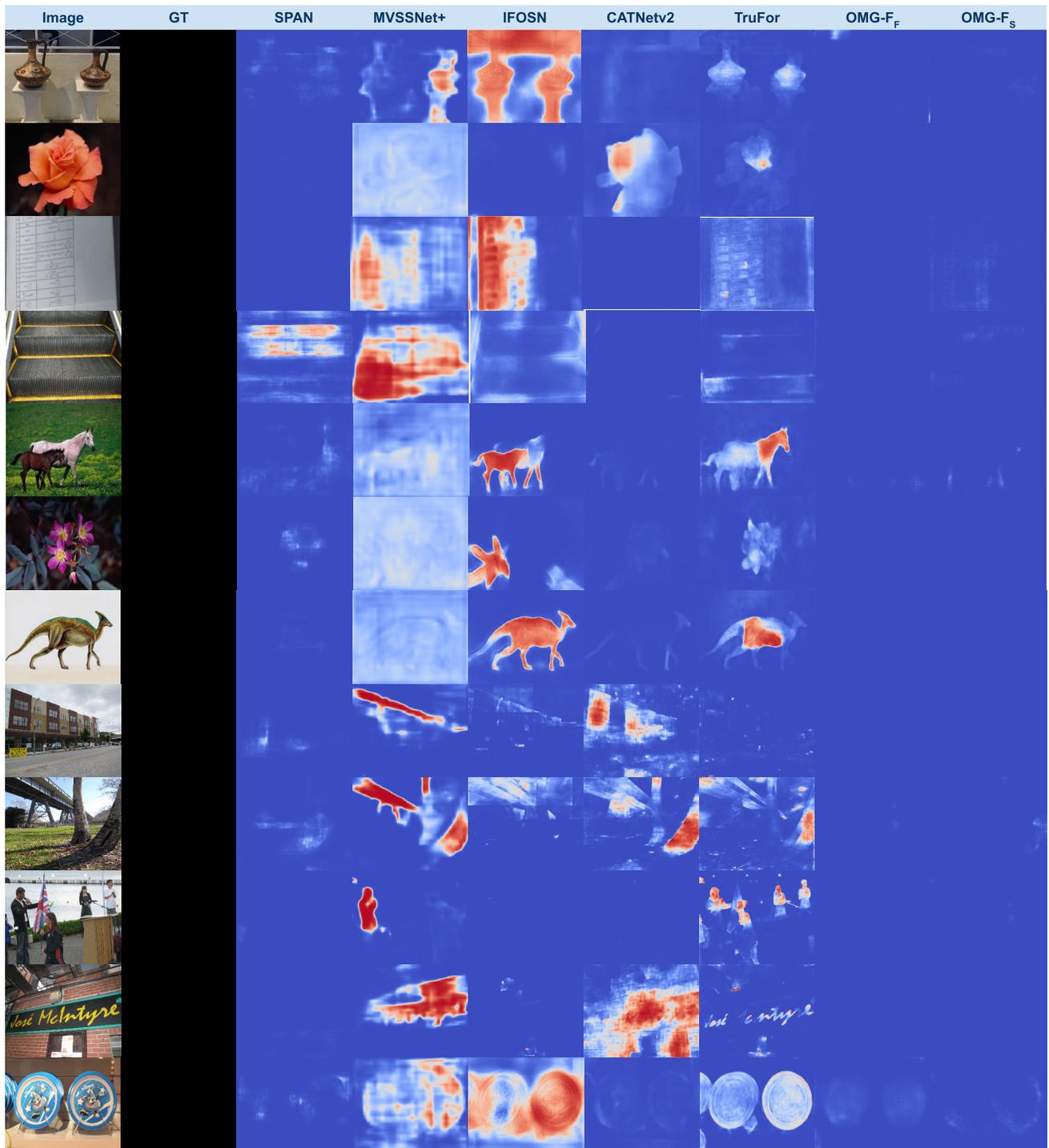}
  \caption{Additional qualitative evaluation results on authentic images. From left to right, the image in question, the ground truth mask, the outputs of five recent methods, and the outputs of the two variants of our architecture are displayed.}
  \label{fig:extended_qual_eval_auth}
\end{figure*}

\textbf{Additional qualitative evaluation:} Finally, we present an extensive qualitative evaluation of our score- and feature-level fusion implementations with several forged and authentic samples in \cref{fig:extended_qual_eval_forged} and \cref{fig:extended_qual_eval_auth}, respectively. In forged samples, our models greatly improve the localization mask, while in authentic samples, they considerably decrease the false positives. 

\section{Explainability}

To better understand which parts of the input signals contribute the most to the fused tokens $\bar{z}^{ft}$ (eq. 5), we employ the Grad-CAM \cite{selvaraju2017grad} method. In particular, we compute the gradients of the fused tokens with respect to the $N+1$ different inputs to the TFT in $z$ (eq. 4) in order to isolate the token fusion process and determine from which tokens the information propagates to the next stages. We compute the gradients based on the output of the TFT, using the squared $\ell^2$-norm of the $\bar{z}^{ft}$. In these experiments, we employ the OMG-Fuser$_S$ variant of our architecture due to the easier interpretation of the input signals. The explainability maps for three samples are presented in \cref{fig:explainability}. Our architecture has learned to attend to the correctly estimated regions in the input signals based on the ground truth while ignoring the regions of the input signals containing erroneous predictions. Also, our network focuses on the signals that better capture the forged and the authentic regions separately, \eg on MVSSNet++ for the detection of authentic regions, while exploiting information from the image to resolve ambiguity in the input signals. 

\section{Discussion on Research Ethics}

Our primary ethical consideration while carrying out this research has been the potential for misuse of the proposed method. In particular, as with any image forensics method, the outputs of forgery localization and detection may be misinterpreted by non-experts or misused by malicious actors in an effort to discredit online digital media as being ``manipulated''. This is especially true for methods that result in high false positive rates. Given that OMG-Fuser exhibits consistent improvements in detection accuracy with the integration of additional input forensic signals and demonstrates very low false positive rates, we expect the risk of misuse to be negligible, while at the same time, it enhances the current capabilities of detecting forged online content aimed at spreading disinformation.

\end{document}